\documentclass[journal]{IEEEtran}

\ifCLASSINFOpdf

\else

\fi

\usepackage{stmaryrd}
\usepackage{bm}
\usepackage{amssymb}
\usepackage{booktabs}
\usepackage{graphicx}
\usepackage{float}
\usepackage{graphicx}
\usepackage{stfloats}
\usepackage{amsmath}
\usepackage{epstopdf}
\usepackage{algorithm,algorithmic}
\usepackage{multirow}
\usepackage{bbm}
\usepackage{framed}
\usepackage[colorlinks,linkcolor=red,anchorcolor=green,citecolor=green]{hyperref}
\usepackage[numbers,sort&compress]{natbib}
\usepackage{amsthm}
\usepackage{diagbox}
\usepackage{microtype}
\usepackage{graphicx}
\usepackage{subfigure}
\usepackage{booktabs} 
\usepackage{multicol}
\usepackage{lipsum}

\newcommand{\Cmat}{{\bf C}}
\newcommand{\Dmat}{{\bf D}}

\newcommand{\Hmat}[0]{{{\bf H}}}

\newcommand{\Xmat}{{\bf X}}
\newcommand{\Ymat}[0]{{{\bf Y}}}
\newcommand{\Zmat}{{\bf Z}}

\newcommand{\nv}{\boldsymbol{n}}

\newcommand{\xv}{\boldsymbol{x}}
\newcommand{\yv}{\boldsymbol{y}}

\newcommand{\zv}{\boldsymbol{z}}

\newcommand{\ts}{^{\top}}
\newcommand{\ie}{{\em i.e.}}

\begin{document}
%
\title{Reinforcement Learning for \\Adaptive Video Compressive Sensing}
%
%
%

\author{Sidi Lu,
        Xin Yuan,~\IEEEmembership{Senior Member,~IEEE},
        Aggelos K Katsaggelos,~\IEEEmembership{Fellow,~IEEE}
        and~Weisong Shi,~\IEEEmembership{Fellow,~IEEE}
\IEEEcompsocitemizethanks{
\IEEEcompsocthanksitem Sidi Lu and Weisong Shi are with Department of Computer Science, Wayne State University, Detroit, MI 48202, USA. E-mail: \{lu.sidi,weisong\}@wayne.edu.
\IEEEcompsocthanksitem  Xin Yuan is with Nokia Bell Labs, 600 Mountain Ave., Murray Hill, NJ 07974, USA. E-mail: xyuan@bell-labs.com.
\IEEEcompsocthanksitem  Aggelos K. Katsaggelos is with Department of Electrical and Computer Engineering, Northwestern University, Evanston, IL 60208, USA, E-mail: aggk@eecs.northwestern.edu.
}
}

\maketitle

\begin{abstract}
We apply reinforcement learning to video compressive sensing to adapt the compression ratio. Specifically, video snapshot compressive imaging (SCI), which captures high-speed video using a low-speed camera is considered in this work, in which multiple ($B$) video frames can be reconstructed from a snapshot measurement.  One research gap in previous studies is how to adapt $B$ in the video SCI system for different scenes. In this paper, we fill this gap utilizing reinforcement learning (RL). An RL model, as well as various convolutional neural networks for reconstruction, are learned to achieve adaptive sensing of video SCI systems. Furthermore, the performance of an object detection network using directly the video SCI measurements {\em without reconstruction} is also used to perform RL-based adaptive video compressive sensing. Our proposed adaptive SCI method can thus be implemented in low cost and real time.
Our work takes the technology one step further towards real applications of video SCI.
\end{abstract}


%
\IEEEpeerreviewmaketitle

\section{Introduction}
\label{submission}
\IEEEPARstart{V}{ideo compressive sensing} is a promising technique inspired by compressive sensing (CS)~\cite{Donoho06_CS,Candes06_Robust}. 
We consider the snapshot compressive imaging (SCI)~\cite{Yuan2021_SPM,Jalali19TIT_SCI,Liu19_PAMI_DeSCI}, which uses a two-dimensional (2D) detector to sample the high-dimensional data, such as high-speed video~\cite{Llull13_OE_CACTI} and hyperspectral images~\cite{Meng20ECCV_TSAnet}. 
In particular, we focus on the video SCI system, which is  representative of both video CS and SCI. 
The underlying principle of video SCI is to modulate the high-speed video with a higher frequency than the sampling rate of the camera~\cite{Llull13_OE_CACTI,Reddy11_CVPR_P2C2,Hitomi11_ICCV_videoCS}. In this manner, video SCI can utilize a low-speed camera to capture high-speed videos. 
Most recently, by using deep learning (DL) algorithms~\cite{Cheng20ECCV_Birnat,Cheng2021_CVPR_ReverSCI,Wang2021_CVPR_MetaSCI} for real-time reconstruction, end-to-end sampling and reconstruction video SCI systems have been built~\cite{Qiao2020_APLP}. 
Now it is the right time to take the developments one step further and make SCI systems suitable for real applications. 

Bearing this concern in mind, this paper considers the video SCI system from the perspective of adaptive sensing~\cite{Yuan13ICIP}. This is motivated by real applications, as scenes are dynamic, of various backgrounds and speeds and thus different compression ratios should be used. 
Moreover, the compression ratio should be {\em adaptively} adjusted for different scenes or as the contents in the scene change.  
In this paper, we address this challenge by {\em reinforcement learning} (RL)~\cite{Sutton1998}.
Specifically, we treat the video SCI system as an {\em agent} and the scene being captured as the {\em environment}. By developing the {\em policy} and {\em reward}, we build an end-to-end RL-based adaptive video SCI system.

\subsection{Video Compressive Sensing}

\begin{figure*}[t]
\centering
\includegraphics[width = \textwidth]{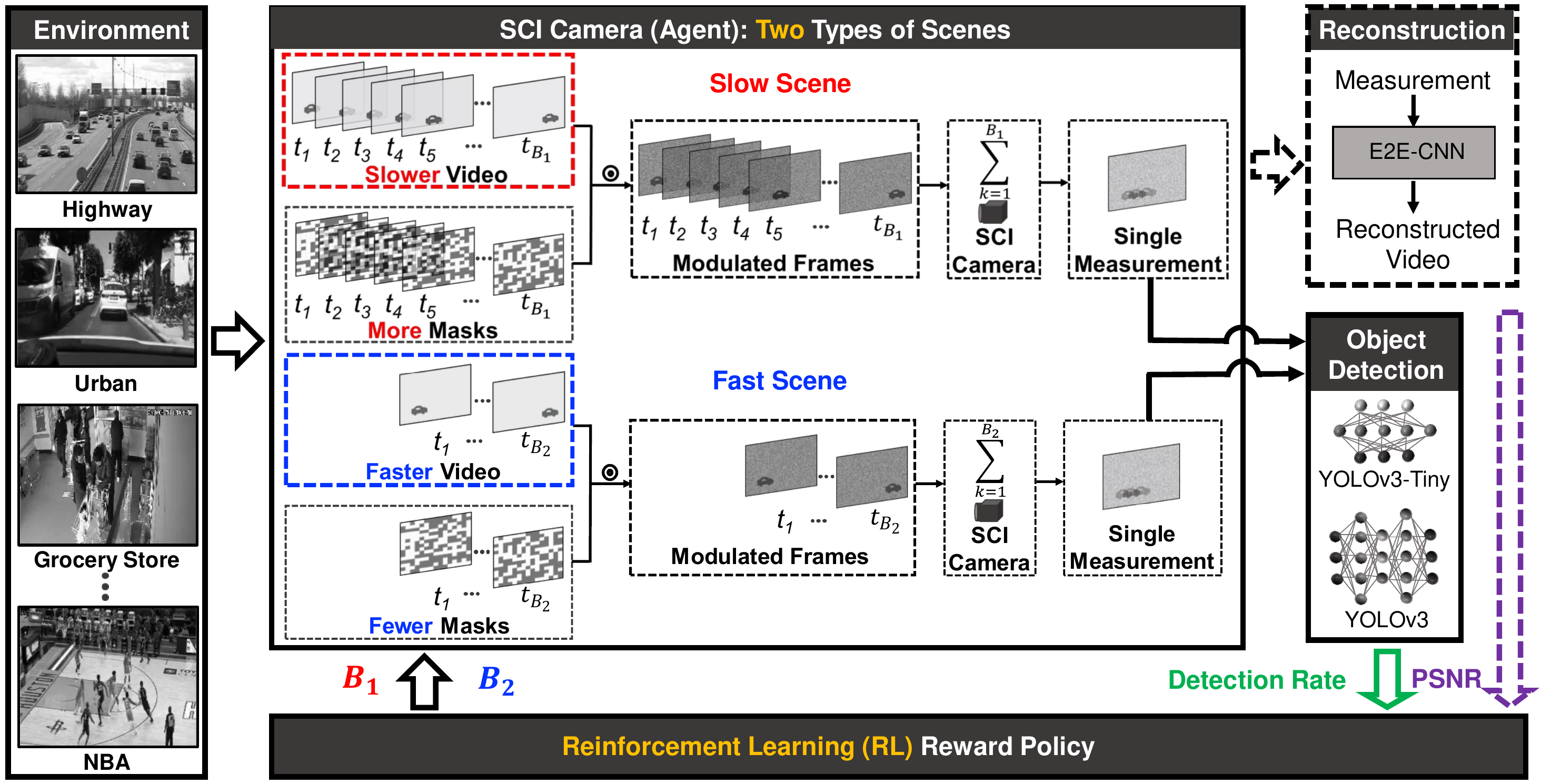}
\caption{The framework of video Snapshot Compressive Imaging (SCI) and reinforcement learning (RL) for temporal adaptive sensing. 
SCI cameras (middle-top) are used to capture and thus sense the environment (left) with an adaptive compression ratio ($B$) determined by the RL policy (bottom-middle). In SCI, every $B$ frames are {\em compressed} to a single measurement, which is sent to the {\em object detection} module (bottom-right, YOLOv3 \cite{redmon2018yolov3} and YOLOv3-Tiny \cite{xiao2019target} are used here) directly; optionally, the measurement can also be sent to the {\em reconstruction} module (top-right) to perform video recovery. The end-to-end convolutional neural network, named E2E-CNN \cite{Qiao2020_APLP}, is used to reconstruct high-speed video frames from a single measurement. The detection rate and optionally the PSNR of the reconstructed video (available during training) are sent to the RL module to adjust $B$ for different scenes.
Here, $\odot$ denotes the element-wise product. $B_1$ is large for a slow motion scenes while $B_2$ is small for a high-speed motion scenes. Note that only one $B$ ($B_1$ or $B_2$ as in the two examples) is the output of the RL module at each time step. {Only {\em a single} SCI camera is used, instead of two camera agents, to capture slower and faster scenes. The goal of distinguishing slower and faster parts of a scene in this plot is to highlight that our work can {\em adapt the compression ratio ($B$)}  by the proposed RL policy.}}
\label{fig:SCIRL}
\end{figure*}

As depicted in Fig.~\ref{fig:SCIRL} (top-middle), for a high-speed video with $B$ frames $\Xmat\in {\mathbb R}^{N_x\times N_y \times B}$, a different mask (coding pattern) $\Cmat \in {\mathbb R}^{N_x\times N_y \times B}$ is imposed on each of them, and then these modulated frames are summed into a single measurement $\Ymat\in {\mathbb R}^{N_x\times N_y}$, and we define $B$ as the compression ratio. This process can be recognized as a {\em hardware encoder} and the key ingredient is the high-speed modulation. Different approaches have been proposed in the literature, such as a shifting mask~\cite{Llull13_OE_CACTI, koller2015high} or a digital micromirror device~\cite{Reddy11_CVPR_P2C2,Sun17OE}, to achieve this modulation.  

The other important part of video SCI is the {\em software decoder}, or the inverse algorithms, to reconstruct the high-speed video from the compressed measurement given the masks~\cite{Yuan2021_SPM}. For a long time, the reconstruction algorithm was the bottleneck precluding the wide applications of video SCI. In the literature, diverse optimization methods developed for CS have been used~\cite{Bioucas-Dias2007TwIST,Yuan16ICIP_GAP,Yang14GMM,Yang14GMMonline,Yang20_TIP_SeSCI}. 
It is only in the last few years that the quality of the reconstructed videos has been significantly improved and they can be used in our daily life~\cite{Liu19_PAMI_DeSCI}. One common drawback of these model-based optimization methods is the slow reconstruction speed. Most recently, this drawback has been ameliorated by DL neural networks~\cite{Cheng20ECCV_Birnat,Qiao2020_APLP,Ma19ICCV,Iliadis_DSP2020}, and has led to high-speed high-quality reconstructions. In short, the hardware encoder and DL based software decoder have now paved the way of end-to-end video SCI systems to be used in our daily life~\cite{Lu20SEC}.
 
\subsection{Temporal Adaptive Sensing in Video SCI}
From the application perspective, to deploy video SCI systems into our daily life, different settings are required for different scenes.
Taking video surveillance as an example, video SCI cameras can significantly reduce memory and transmission bandwidth costs, while, recovering the high-speed video if needed.  
However, a fixed compression ratio ($B$:1) is clearly not optimal in this case, since when there are no moving objects in the scene, a large $B$ can be used, while when a high-speed object exists in the scene, a small $B$ is desired for maintaining high quality reconstruction (Fig. \ref{fig:SCIRL}). 
Moreover, we expect that the video SCI system can adjust this $B$ value automatically.
This is what we refer to as {\em temporal adaptive sensing}\footnote{The other proposal to adaptive sensing is to adjust the compression ratio spatially as a function of content (different places on the image plane). However, this will pose a significant challenge for the hardware design and thus we do not consider it here.} and we aim to address it by RL (Fig. \ref{fig:SCIRL} bottom) in this paper.

\subsection{Related Work}
Although the idea of adaptive CS has been proposed for a long time, in most cases it applies to spatial CS, \ie, following the single pixel camera architecture~\cite{Duarte08SPM}. 
By contrast, for adaptive sensing in video CS considered in this paper, only a few papers exist and the one closely related to ours is~\cite{Yuan13ICIP}, which considers the same problem but by using a motion estimation method to adapt $B$. 
However, both the reconstruction algorithm and the adaptive sensing framework developed therein produce low quality results. During the past eight years, the reconstruction algorithms of video CS have been improved significantly, especially the ones based on DL~\cite{Qiao2020_APLP,Yuan2020_CVPR_PnP,Cheng20ECCV_Birnat}. 
Moreover, the look-up table used in \cite{Yuan13ICIP} only connected adaptive temporal sensing with motion estimation and did not consider the scene complexity and object detection rate, which are important factors for the adaptive framework developed in this paper.

\subsection{Reinforcement Learning}


Reinforcement learning \cite{Sutton1998,kiumarsi2017optimal, moerland2020model} is an online algorithm designed to optimize behavioral strategies in sequential decision problems \cite{liu2020weighing}, wherein agents continuously interact with unknown environments and seek behavioral policies to maximize the expected cumulative reward. Many challenging benchmark tasks can be performed in this framework, such as robotics \cite{lillicrap2015continuous, levine2016end}, high-dimensional continuous control simulations \cite{schulman2017proximal, lillicrap2015continuous}, the game of Go \cite{silver2016mastering}, Atari \cite{mnih2013playing}, and competitive video games \cite{vinyals2017new, silva2017moba}.
An RL agent uses a policy to control its behavior, where the policy is a mapping from obtained inputs to actions. 
One main difference between RL and supervised learning is that the RL agent is never told the optimal action, instead it receives an evaluation signal indicating the goodness of the selected action.
This matches well with an adaptive video CS considered in this work, where the SCI camera usually does not know the environment and the objects in the scene being captured are dynamic and their speed can vary over time.



\subsection{Contributions of This Paper}
In this work, we revisit the temporal adaptive sensing problem in video CS by using three new modules: 
$i$) end-to-end convolutional neural network (E2E-CNN) \cite{Qiao2020_APLP} based reconstruction, $ii$) RL for adaptive sensing control, and $iii$) edge compression based applications~\cite{Lu20SEC} by conducting object detection directly on the video SCI measurements without reconstruction.
Our new regime brings video SCI closer to real applications, such as, connected and autonomous vehicles.

{Remarkably, previous work \cite{Lu20SEC} proved that the object detection accuracy utilizing the compressed measurements (without reconstructing the high-speed video) is  close  to  the one obtained using the original video. Therefore, the advantage of using SCI is clear since it accelerates inference by performing {\em measurement-based object detection}. However, there is a non-negligible trade-off between detection accuracy, reconstruction quality, and compression ratio, which hinders the real applications of measurement-based object detection across diverse fields significantly. In this context, the core innovation of our study is to provide actionable insights into solving a real application challenge by automatically determining the optical compression ratio using RL, which can accelerate the deployment of SCI cameras and measurement-based object detection for time-sensitive applications.} 

The rest of this paper is organized as follows. Section~\ref{Sec:model} describes the proposed RL model for adaptive video CS. Extensive results are presented in Section~\ref{Sec:results} and Section~\ref{Sec:conclusion} concludes the paper.

\section{Proposed RL Model for Adaptive Video CS \label{Sec:model}}
In this section, we first describe the mathematical model of video SCI and briefly introduce the state-of-the-art deep learning based reconstruction approaches. The proposed RL based adaptive sensing is detailed in Sec.~\ref{Sec:RL_aaptive}.  

\subsection{Mathematical Model of Video SCI}
Following Fig.~\ref{fig:SCIRL} (top-middle), a $B$-frame dynamic scene $ \Xmat \in \mathbb{R}^{N_x \times N_y \times B}$  is modulated by $B$ fast updated masks $\Cmat\in \mathbb{R}^{N_x \times N_y \times B}$, and then the modulated video frames are integrated into a single measurement frame  $\Ymat \in \mathbb{R}^{N_x\times N_y}$ by a camera sensor with the exposure time of these $B$ frames. This process can be expressed as
\begin{equation}\label{Eq:System}
  \Ymat =  \sum_{b=1}^B \Cmat_b\odot \Xmat_b + \Zmat,
\end{equation}
where $\Zmat \in \mathbb{R}^{N_x \times N_y }$ denotes noise, $\Cmat_b = \Cmat(:,:,b)$ and $\Xmat_b = \Xmat(:,:,b) \in \mathbb{R}^{N_x \times N_y}$ the $b$-th mask and the corresponding video frame, and $\odot$ the Hadamard (element-wise) product. 
Using a vectoring operator, define $\yv = \text{Vec}(\Ymat) \in \mathbb{R}^{N_x N_y}$ and $\zv= \text{Vec}(\Zmat) \in \mathbb{R}^{N_x N_y}$. Similarly,  define  $\xv \in \mathbb{R}^{N_x \times N_y \times B}$ as
\begin{equation}
 \xv = \text{Vec}(\Xmat) = [\text{Vec}(\Xmat_1)\ts,..., \text{Vec}(\Xmat_B)\ts]\ts.
\end{equation}
The measurement process in \eqref{Eq:System} can thus be expressed  as 
\begin{equation}\label{Eq:ghf}
 \yv = [\Dmat_1,...,\Dmat_B] \xv + \zv,
\end{equation}
where, $\Dmat_b = \text{diag}(\text{Vec}(\Cmat_b)) \in {\mathbb R}^{N \times N}$, for $b =1,\dots B$ and $N = N_x N_y$. The sensing matrix $\Hmat=[\Dmat_1,...,\Dmat_B] \in \mathbb{R}^{N\times N B}$ in video SCI is  highly structured  and sparse. 
It has been shown in~\cite{Jalali19TIT_SCI} that, if the signal is structured enough,  there exist SCI recovery algorithms with bounded reconstruction error for $B>1$.

\subsection{Deep Learning for Reconstruction and Detection}
Reconstruction aims to recover high quality videos from the compressed measurement $\Ymat$ captured by the SCI camera. Significant efforts have been made to develop new reconstruction algorithms in the past decade since high quality videos were recognized as the main output of a SCI camera. Recently, with the aid of DL, this challenge has been addressed using deep convolutional neural networks (CNN) and recurrent neural networks (RNN)~\cite{Cheng20ECCV_Birnat,Qiao2020_APLP}. 
Most recently, motivated by the demanding application of connected and autonomous vehicles, an SCI-vehicle-edge-cloud framework has been proposed~\cite{Lu20SEC}. This leads us to think deeper about the main objective of an SCI camera. In addition to the high quality videos, which is of course very important for the subsequent processing, we also need {\em fast detection and real-time control}, from the raw measurements if possible. 
Studies in~\cite{Lu20SEC} have proved that this dual objective is feasible and thus demonstrated the promising applications of SCI. 

Taking one step further, it is not optimal to use a fixed compression ratio ($B$:1) in SCI cameras due to the dynamic nature of the scene. This dictates the research on adaptive sensing and in this paper, we fill this gap by RL since an SCI camera itself is an agent to sense (and thus capture) the environment. 

\subsection{RL for Adaptive Sensing~\label{Sec:RL_aaptive}}
In RL, the goal of the agent is formalized with respect to a specific signal passing from the environment to the agent. This signal is referred to as the reward ($r$), which is a simple number at each time step ($t$), \ie,  $r_{t}\in \mathbb{R}$. To be specific, the goal of this work is to maximize the cumulative reward that the agent (SCI camera) receives.

\subsubsection{States and Transition Graph}
To make the SCI camera learn to automatically determine the optimal $B$, we have provided a reward at each time step corresponding to the SCI camera's forward action $a$ including increasing $B$, keeping the current value of $B$, or decreasing $B$. More specifically, in this work, we assume that six reconstruction models (E2E-CNN) with different values of $B$, \ie, $B$ = \{6, 8, 10, 12, 15, 20\} have been trained for real-world applications, comprising a \texttt{state set} $\cal S$ = \{6, 8, 10, 12, 15, 20\}. These values are heuristically selected by extensive experiments on various videos to be able to obtain decent reconstructions.

At each state, the SCI camera can decide whether to $i)$ actively increase $B$, $ii)$ keep the current value of $B$, or $iii)$ decrease $B$. 
Note that `increase' and `decrease' can skip intermediate values of $B$; for example, our policy allows changing $B=15$ to $B=6$ as in real life applications, when a red traffic light or an accident can suddenly halt all cars (a large $B$ can be used) while all cars will speed up (a small $B$ is required)  when the traffic light turns green. 
We use ${a}$ to represent the \texttt{action set} and ${ a} = \left \{ decrease, keep, increase \right \}$, which is predicted by the RL model. ${S}'$ indicates the updated state after conducting each $a$. As to each action step, RL provides the corresponding \texttt{reward} $r \left ( S,a, {S}'\right )$, which is related to the corresponding environment.


\begin{table}[h]
\centering
\caption{State transition table of the proposed RL for adaptive video CS by only considering three states, \ie, $S = 6, 10, 15$. $increase^{2}$ denotes the increase of $B$ skips from 6 to 15 and similarly $decrease^{2}$ denotes the decrease from 15 to 6 with $\alpha,\beta\in[0,1]$. }

\begin{tabular}{ccc|cl}
$S$ & $a$ & ${S}'$ & $p\left ({S}'\mid S,a  \right )$ & $r \left ( S,a, {S}'\right )$ \\ \hline\hline
6 & decrease & 6 & 0 & $-$ \\
6 & keep & 6 & 1 & $r_{keep}$ \\
6 & increase & 10 & $\alpha$ & $r_{increase}$ \\
6 & $increase^{2}$ & 15 & $1-\alpha$ & $r_{increase}$ \\
10 & decrease & 6 &  1 & $r_{decrease}$ \\
10 & keep & 10 & 1 & $r_{keep}$\\
10 & increase & 15 & 1 & $r_{increase}$\\
15 & $decrease^{2}$ & 6 & $\beta$ & $r_{decrease}$ \\
15 &  decrease & 10 & $1-\beta$  & $r_{decrease}$ \\
15 & keep & 15 & 1 & $r_{keep}$\\
15 & increase & 15 & 0 & $-$\\
\hline\hline
\end{tabular}
\label{tab:transition}
\end{table}

\begin{figure}[h]
\centering
\includegraphics[scale=0.5]{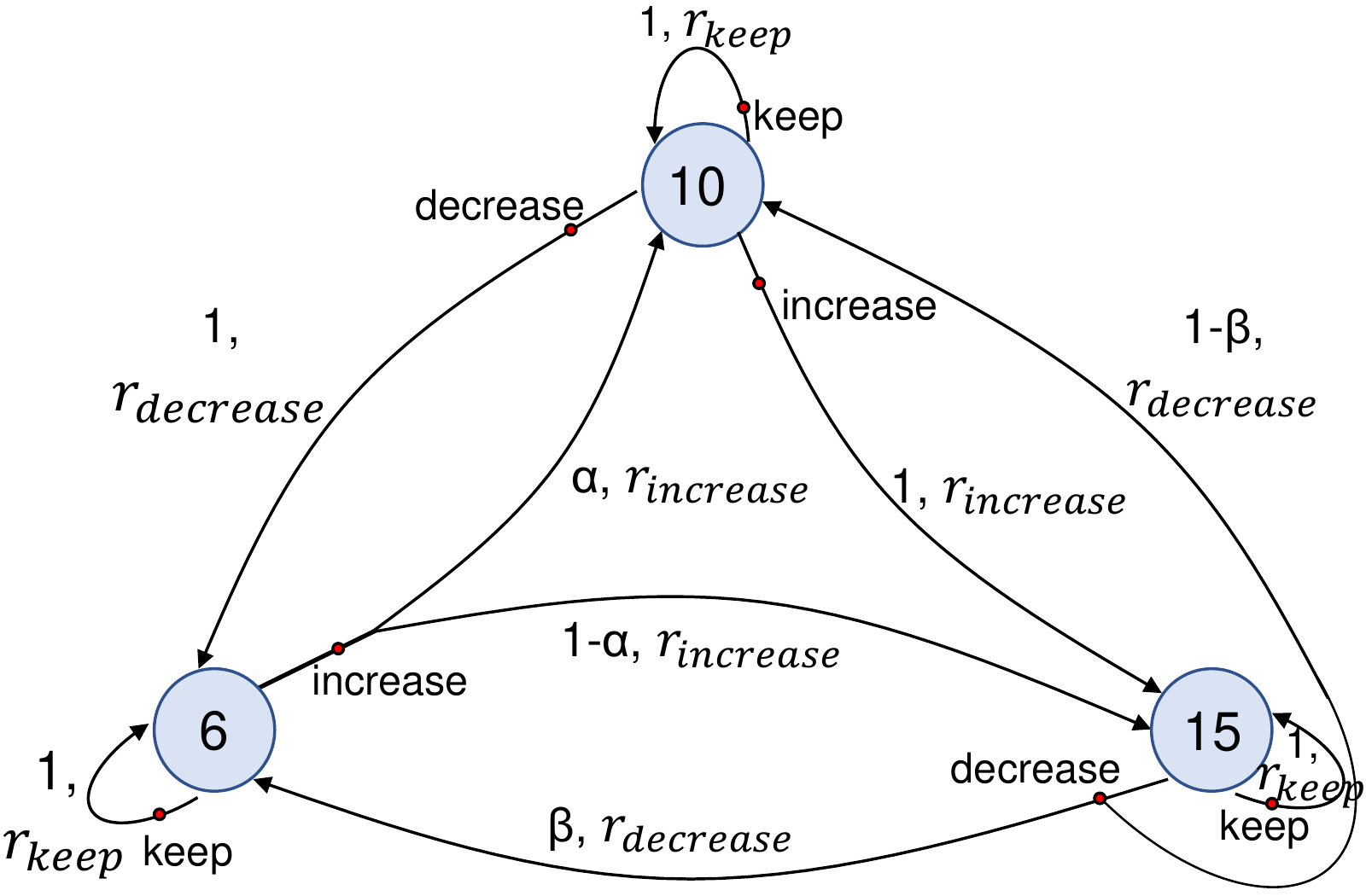}
\caption{State transition graph of Table~\ref{tab:transition}.}
\label{fig:trans}
\end{figure}

Table~\ref{tab:transition} summarizes the dynamics of the transition table for a simple example. 
For the sake of conciseness and concreteness, Table~\ref{tab:transition} only considers three states, \ie, ${\cal S} = \{6, 10, 15\}$. In this example, a period of search that begins with $S = 6$ cannot leave for the new state ${S}'$ with $a = decrease$ since 6 is already the minimum value of $B$; therefore, the corresponding conditional probability $p\left ({S}'\mid S,a  \right ) = 0$ and no related reward $r$ exists. However, with the action of increase, \ie, $S = 6$ and $a = increase$, $S$ could be increased to 10 or 15 (\ie, ${S}'$ = 10 or 15) with probability $\alpha$ and $1-\alpha$, respectively, where $\alpha\in[0,1]$. Similarly, a period of searching undertaken when $S = 15$ and $a = decrease$ ends at ${S}' = 6$ with probability $\beta$ and ${S}' = 10$ with probability $1-\beta$, with $\beta\in[0,1]$. 
The corresponding  state transition graph is shown in Fig.~\ref{fig:trans}. 

\subsubsection{Reward Policy} 
In real-world applications, the reward policy design of RL is highly correlated with the involved deep learning models and the specific scenes. As shown in Fig.~\ref{fig:SCIRL}, the SCI captured measurements are sent to two modules \ie, the \textit{detection} module and \textit{reconstruction} module, to perform object detection and optionally the video reconstruction, respectively. Therefore, we consider the detection rate and PSNR of the reconstructed video as the key performance metrics for the RL module to adjust $B$ for different scenes.

Note that the PSNR can only be used during training as in real applications, no ground truth is available to calculate it. Here, PSNR \cite{poobathy2014edge} refers to the peak-signal-to-noise ratio between two images, and we use it to evaluate the performance of the reconstruction model (E2E-CNN)~\cite{Qiao2020_APLP}. More specifically, let $\Xmat^* \in {\mathbb R}^{N_x\times N_y \times B \times G}$ denote the ground truth video group, where $G$ denotes the number of measurements being used, and $\hat \Xmat$ be the reconstructed video by the E2E-CNN with the same size as $\Xmat^*$. The average PSNR of the video group is given by:
\begin{equation}
\footnotesize
    {\rm PSNR} = \frac{1}{BG}\left[ -10 \log \frac{\sum_{n_x=1}^{N_x} \sum_{n_y = 1}^{N_y} (\hat{x}_{n_x, n_y,b,g} -{x}^*_{n_x, n_y,b,g} )^2}{N_x N_y} \right]
\end{equation}
where $\hat{x}_{n_x, n_y,b,g}$  and ${x}^*_{n_x, n_y,b,g}$ denote the $(n_x, n_y)$-th pixel in the $b$-th frame of the $g$-th measurement in the estimated video and ground truth video, respectively. Usually, the lower the value of $B$, the higher the PSNR (smaller error), and the better the quality of the reconstructed image.

{In this work, the goal of video CS is to conduct object detection on the {\em measurements} (compressed data captured by SCI cameras) with an adaptive compression ratio ($B$). Therefore, apart from PSNR, the detection rate is a good objective metric for this task to assist the adjustment of $B$, \ie, it is also sent to the RL module to adjust $B$ for different scenes. 
Other metrics can also be used in the future for the same or different tasks. }



%


		

\setlength{\intextsep}{0pt} 
\begin{algorithm}[tb]
	\caption{RL for Adaptive video CS}
	\begin{algorithmic}[1]
		\REQUIRE$\Hmat$, detection model (and reconstruction models).
		\STATE Initial $B$, $drth$ as the threshold of acceptable detection rate, and optionally $psnrth$ as the threshold of acceptable PSNR.
		\WHILE{Capturing}
		\STATE Capture measurement of $\Ymat$.
		\STATE Perform detection on the measurement and output the detection rate. {\em Optionally} conduct the reconstruction and calculate PSNR during training.
		\STATE RL policy update by detection rate (and PSNR).
     \begin{multicols}{2}
		\IF{$detect\_rate$\textless $drth$~~\\}
             \IF {$a$=decrease OR\\ ($a$=keep AND~~\\ $B$=$B_{min}$)}
                \STATE $r\gets r_1$
             \ELSE
                \STATE $r\gets r_2$
             \ENDIF
        \ELSE 
            \IF {$a$=increase OR\\ ($a$=keep AND ~~\\$B$=$B_{max}$)}
                \STATE $r\gets r_1$
             \ELSE
                \STATE $r\gets r_2$
             \ENDIF
        \ENDIF
        \\~
        \IF {PSNR provided}
        \IF {PSNR $>$ $psnrth$}
            \IF {$r$ $>$ 0}
                \STATE $r\gets r*\lambda_1$
            \ELSE
                \STATE $r\gets r*\lambda_2$
            \ENDIF
        \ELSE
            \IF {$r$ $>$ 0}
                \STATE $r\gets r*\lambda_2$
            \ELSE
                \STATE $r\gets r*\lambda_1$
            \ENDIF
        \ENDIF
        \ENDIF
        \end{multicols}
		\STATE Output $B$, $r$.
		\ENDWHILE
	\end{algorithmic}
	\label{algo:RL_SCI}
\end{algorithm}

Algorithm~\ref{algo:RL_SCI} presents the RL reward mechanism for adaptive temporal video CS. As depicted in it, after defining the sensing matrix $\Hmat$ and the initial $B$, the RL module will predict the action (\ie, increase the value of $B$, keep the current value, or decrease it) based on the captured measurement of $\Ymat$, and update $B$ accordingly. We will then perform object detection through YOLOv3-Tiny on the measurements and calculate the detection rate. Here, YOLOv3-Tiny \cite{huang2018yolo} is a light-weight DL algorithm designed for resource-constrained devices, with superior advantages on fast object detection due to the significantly reduced parameters. Optionally, the measurements can also be sent to the {reconstruction} module for video recovery, and the PSNR of the reconstructed video (available during training) will be sent to the RL module to adjust $B$ for different scenes.

 \vspace{0.2cm}
\noindent \textbf{Lines 6-11:} The RL module first defines the thresholds (lower bounds) of the acceptable detection rate and PSNR as $drth$ and $psnrth$, respectively. The higher the values of $drth$ and $psnrth$ the smaller the value of $B$. Consider a round of capturing as an example; if the calculated detection rate is smaller than the threshold, \ie, $detect\_rate$ $<$ $drth$, it reveals that the current $B$ is larger than the optimal value, so we expect the RL module to output a smaller $B$. In this context, if $i)$ the corresponding action $a$ indicates to decrease $B$, or $ii)$ the action $a$ is to keep the current $B$ when $B$ already achieves its minimum value, then the RL module will assign a positive reward $r_{1}$ as encouragement; otherwise, it will assign a negative reward $r_{2}$ as penalty.

\vspace{0.2cm}
\noindent \textbf{Lines 12-18:} Similarly, if $detect\_rate$ $>$ $drth$, it reveals that the current $B$ is smaller than the optimal value, so we expect the RL module to output a larger $B$. In this context, if $i)$ the corresponding action $a$ indicates to increase $B$, or $ii)$ the action $a$ is to keep the current $B$ and $B$ already achieves its maximum value, then the RL module will assign a positive reward $r_{1}$ as encouragement; otherwise, it will assign a negative reward $r_{2}$ as penalty.

\vspace{0.2cm}
\noindent \textbf{Lines 19-33:} Optionally, if reconstruction is conducted and the corresponding PSNR is provided, the reward mechanism will take it into account: $i)$ when PSNR $>$ $psnrth$ (\ie, revealing that the RL module should increase $B$), if the current cumulative reward $r$ is positive, the RL module will update the reward by $r\cdot\lambda_1$ ($\lambda_1 \in \left ( 1,2 \right )$) to increase the related reward; otherwise, the reward will be updated by $r\cdot\lambda_2$ ($\lambda_2 \in \left ( 0,1 \right )$) to weaken the reward. $ii)$ When PSNR $<$ $psnrth$ (\ie, revealing that the RL module should decrease $B$), if the current cumulative reward $r$ is positive, the RL module will update the reward by $r\cdot\lambda_2$ to weaken the related reward; otherwise, the reward will be updated by $r\cdot\lambda_1$ to increase the reward. Finally,  Algorithm~\ref{algo:RL_SCI} will output $B$ and the cumulative reward $r$.

Specifically, in our experiments, during training when PSNR is available, we consider three scenarios: $i)$ PSNR$<$24, $ii)$ 24 $\leqslant$ PSNR$\leqslant$28, and $iii)$ PSNR $>$ 28. The range 24 $\leqslant$PSNR$\leqslant$28 indicates a good performance of the reconstruction model. Since we expect to obtain a relatively higher $B$, we set the corresponding reward to $r$ = $\left | {\text {PSNR}} - 24  \right | \cdot B$; this way, a higher $B$  will provide a higher reward, encouraging the agent to figure out a higher $B$ while guaranteeing the reconstruction quality. When PSNR$<$24, which denotes a poor quality reconstruction, we should reduce $B$; therefore, the reward $r$ is negative as a punishment. Similarly, if PSNR$>$28 in the current time step, we could further improve the value of $B$, so the reward $r$ is positive to encourage a higher $B$. Although the specific positive and negative rewards depend on the specific scene, the basic idea is the same.


\section{Evaluation Results \label{Sec:results}}

\subsection{Datasets and Experiment Setting}
We choose four case studies to show how the proposed RL module can automatically adjust $B$ for different scenes, including urban, highway, grocery store, and NBA scenes. For each case study, we select a specific dataset to train and test the RL module.

\vspace{2mm}

\noindent \textbf{Urban Dataset:} We selected the public dataset of traffic video (PDTV) \cite{saunier2014public} which provides traffic videos at three intersections with annotations for real transportation applications, such as tracking road users and detection of pedestrian infractions. The video dataset was collected at three sites of Belarus and Canada with a resolution of 640 $\times$ 480 pixels at 30 frames per second (fps), and the traffic scenes cross diverse traffic, lighting, and weather conditions. 

\vspace{2mm}

\noindent \textbf{Highway Dataset:} The DynTex dataset \cite{peteri2010dyntex} is the first collection of high-quality dynamic texture videos that are structured by videos’ underlying physical processes such as waving motion and discrete units, with the goal of serving as a standard database for dynamic texture research. Nine sequences related to traffic, with a resolution of 720 $\times$ 576 pixels at 30 fps were selected.

\vspace{2mm}
\noindent \textbf{Grocery Store Video Dataset:} These videos are collected from retail surveillance cameras at a middle-sized grocery store. The camera captures top-down views monitoring both the incoming and outgoing customer flow at the entry gate. Eight video clips with a resolution of 1920 $\times$ 1080 pixels at 30 fps were selected.

\vspace{2mm}
\noindent \textbf{NBA Dataset:} This is a publicly available NBA dataset to test our proposed framework on high-speed sport motions. In the video, two groups of basketball players are moving fast, which is significantly different from other scenes. We selected 5 video clips with a resolution of 640 $\times$ 480 pixels at 30 fps for the experiments.

\subsection{Training Details}
\noindent \textbf{E2E-CNN Training and Validation.} We have six compressed versions of the same video sets to train the E2E-CNN reconstruction modules, \ie, using $B$ = $6, 8, 10, 12, 15, 20$ and the network structure proposed in~\cite{Qiao2020_APLP}\footnote{Code from: {\small{\url{https://github.com/mq0829/DL-CACTI}}}.}. We combine the compressed video segments from the selected video datasets for training and testing. We randomly select 80\% of the measurements for training and the rest for validation. Since not all of these public datasets provide annotations, we directly employ the open YOLOv3 network\footnote{The YOLO series algorithms were firstly proposed in \cite{redmon2016you}, and are well known for fast detection speed by simple and clear algorithm structure.
One popular algorithm, YOLOv3 \cite{redmon2018yolov3}, automatically selects the  suitable initial regression frame by incorporating the $K$-means clustering approach for a specific input dataset.} on the original public video dataset to obtain labels (bounding boxes of targets) and treat these labels as the ground truth. 

Following~\cite{Cheng20ECCV_Birnat}, we define the {\em normalized measurement} from the forward model of SCI in~\eqref{Eq:System} as 
\begin{equation}
    \bar{\Ymat} = \frac{\Ymat}{\sum_{b=1}^B \Cmat_b}.
\end{equation}
This normalized measurement removes the mask artifacts especially in the background and we use it to show the speed of the scene when presenting the results.


\vspace{2mm}
\noindent \textbf{RL Training.} 
{The RL algorithm seeks to maximize a certain measure of the agent’s cumulative reward, as the agent interacts with the environment. In this work, we use the OpenAI Gym framework \cite{brockman2016openai} to build the RL environment. OpenAI Gym focuses on the episodic setting of RL, where the agent’s experience is divided into a series of episodes. For each episode, the starting state of the agent is randomly sampled from a distribution, and the interaction proceeds until it reaches a terminal state under the specific environment. For each use case, we selected the related types of video clips to train the RL model on an NVIDIA GPU workstation (4$\times$GeForce RTX 2080 Ti graphics cards), with the goal of maximizing the expectation of total reward per episode, and to achieve a high level of performance in as few episodes as possible.  We {retrained} the object detection model (YOLOv3-Tiny) on the {SCI measurements}, along with the RL model.}



\subsection{Adaptive Sensing Results Based on PSNR}

\begin{figure*}[htbp!]
\centering
\includegraphics[width =\textwidth]{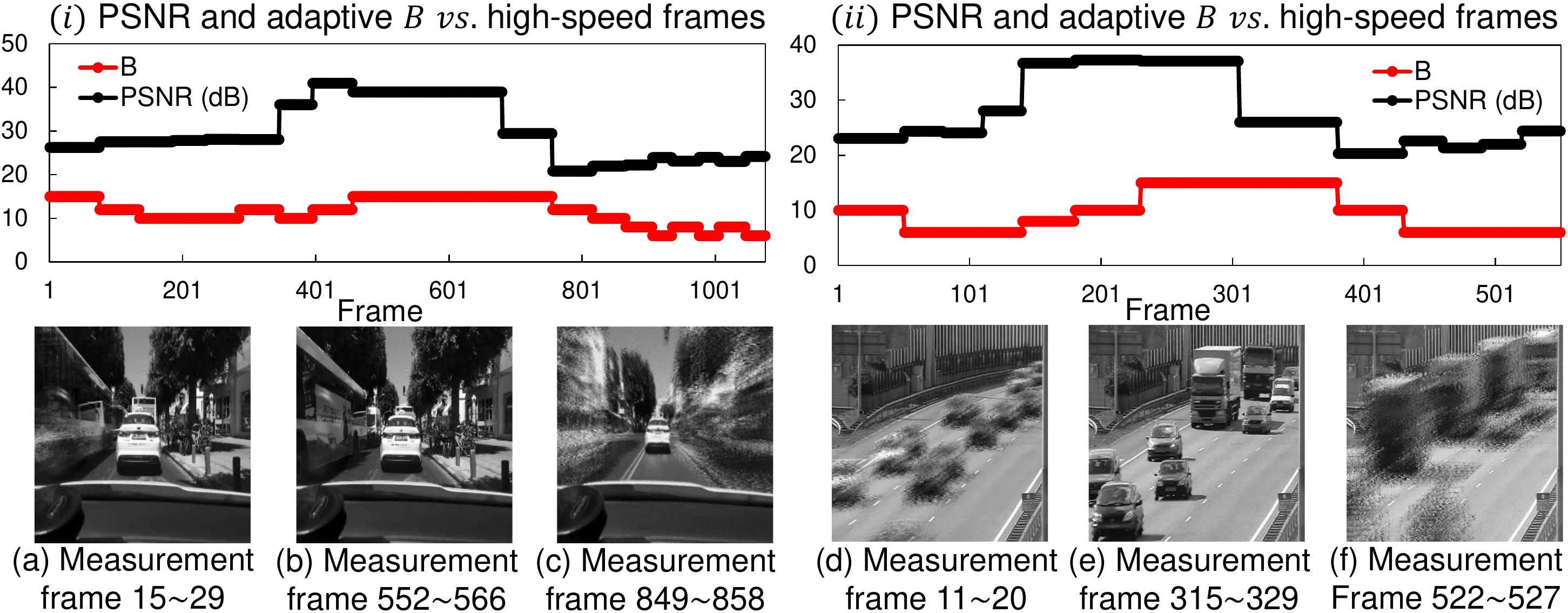}
\caption{($i$-$ii$) Reconstruction PSNR (dB) and adaptive $B$ estimated from the reconstructed Urban (left) and Highway (right) video based on PSNR only, plotted against frame number. (a-f) Normalized measurements with vehicles at different velocities.}
\label{fig:urban}
\end{figure*}

\begin{figure*}[htbp!]
\centering
\includegraphics[width=\textwidth]{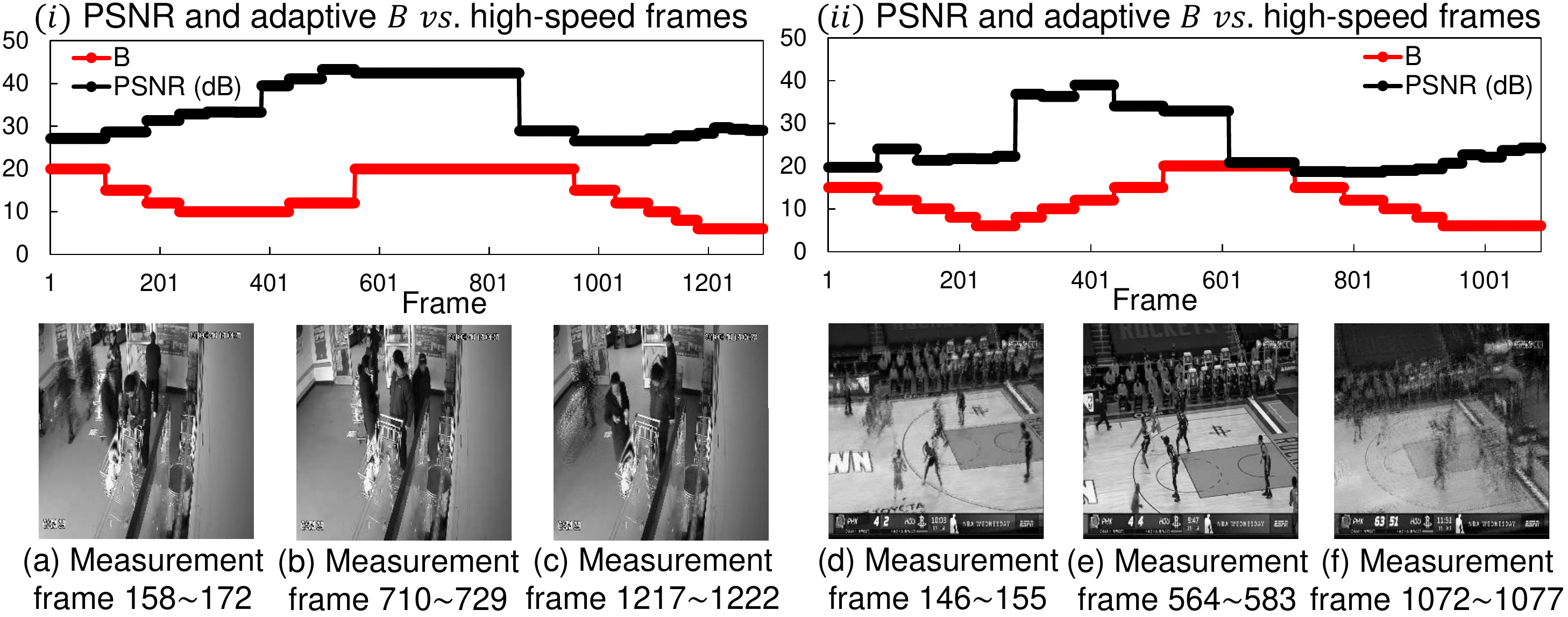}
\caption{($i$-$ii$) Reconstruction PSNR (dB) and adaptive $B$ estimated from the reconstructed Grocery-store video (left) and NBA video (right), all are plotted against frame number. (a-f) Normalized measurements with vehicles at different velocities.}
\label{fig:mart}
\end{figure*}

To prove the concept, we first only consider the reconstruction module with PSNR available but without using the detection rate, aiming to verify the RL module.
The adaptive $B$ results as well as the PSNR are shown in Fig.~\ref{fig:urban} for the Urban and Highway data, and in Fig.~\ref{fig:mart} for the Grocery-store and NBA data.
Note that in the Urban and Highway data, we freeze the videos (in the middle part) and speed them up by skipping frames (last part) to simulate different velocities of the vehicles. 

It can be seen from Fig.~\ref{fig:urban} that  starting from a random $B$, when the video is frozen, RL  will adjust $B$ to a larger value such as 15 and when the video is speeding up in the last hundreds of frames, $B$ is adjusted to a small value such as 6 or 8.
{Differently from these simulated videos, persons in the  grocery store and players in the NBA data change speed by themselves, which are real videos that SCI cameras may be deployed for. Again, as shown in Fig.~\ref{fig:mart}, starting from a random $B$, when the persons or players move fast, our RL module will infer a smaller $B$ and when nobody moves, a large $B$ such as 20 is inferred. When people start to move, $B$ drops again.} These four videos clearly verify that our RL works well with respect to reconstruction quality and PSNR.
The reconstructed video frames can be found in the supplementary material (SM).

Next, we show results based on the detection rate, as the PSNR is usually not available in real cases. 

\begin{figure*}[htbp!]
\centering
\includegraphics[width = \textwidth]{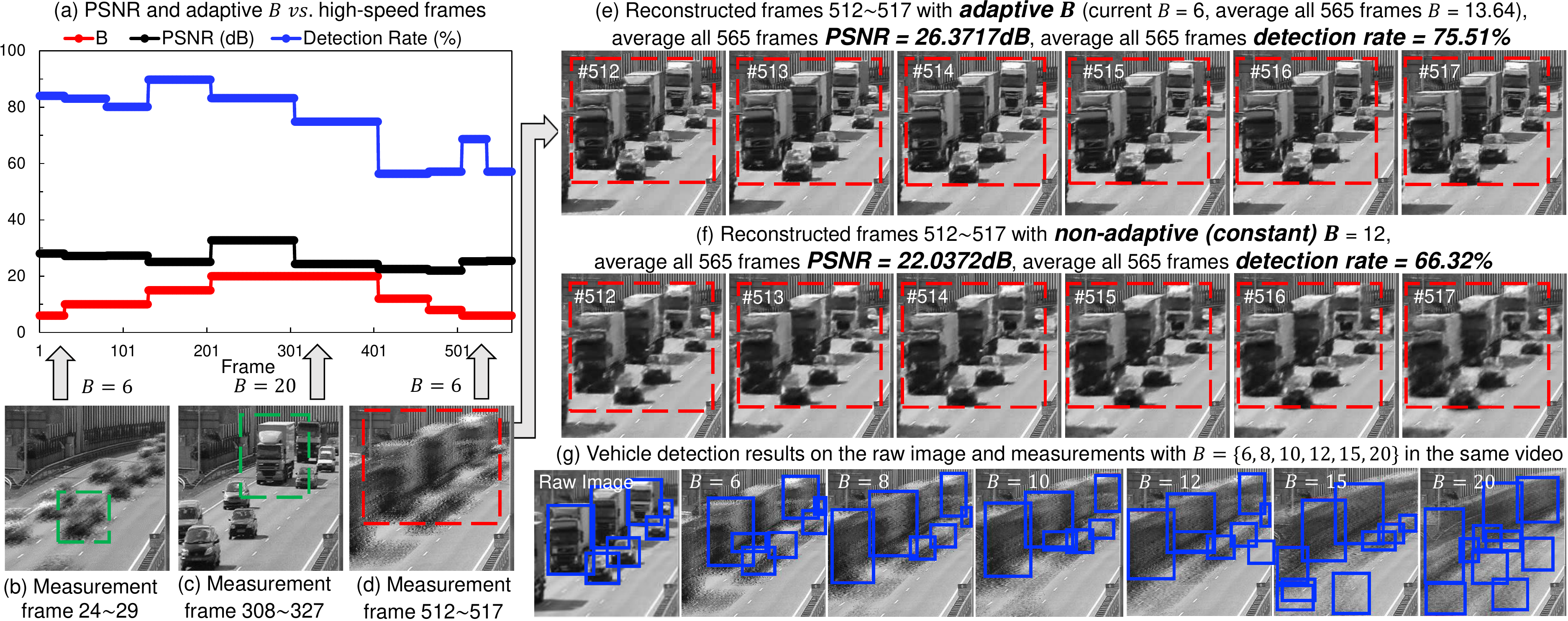}
\caption{Adaptive $B$ from the detection rate on the measurements directly. (a) Reconstruction PSNR (dB) and adaptive $B$ (frames) (average adaptive $B$=13.64) from the {\em measurements}, all are plotted against frame number. (b-d) Normalized measurements when there is no truck, two trucks, and four trucks moving inside the scene, adapted $B$ = 6, 20, 6, respectively. (e) Reconstructed frames 512$\sim$517 from the measurement in (d) with {\em adaptive} $B$. (f) Reconstructed frames 512$\sim$517 with non-adaptive (constant) $B$ = 12. (g) Vehicle detection results on the raw images and measurements with different $B = \{6, 8, 10, 12, 15, 20\}$ in the same video clip. Videos in the SM.}
\label{fig:highwayDetectionBB}
\end{figure*}

\begin{figure*}[htbp!]
\centering
\includegraphics[width=\textwidth]{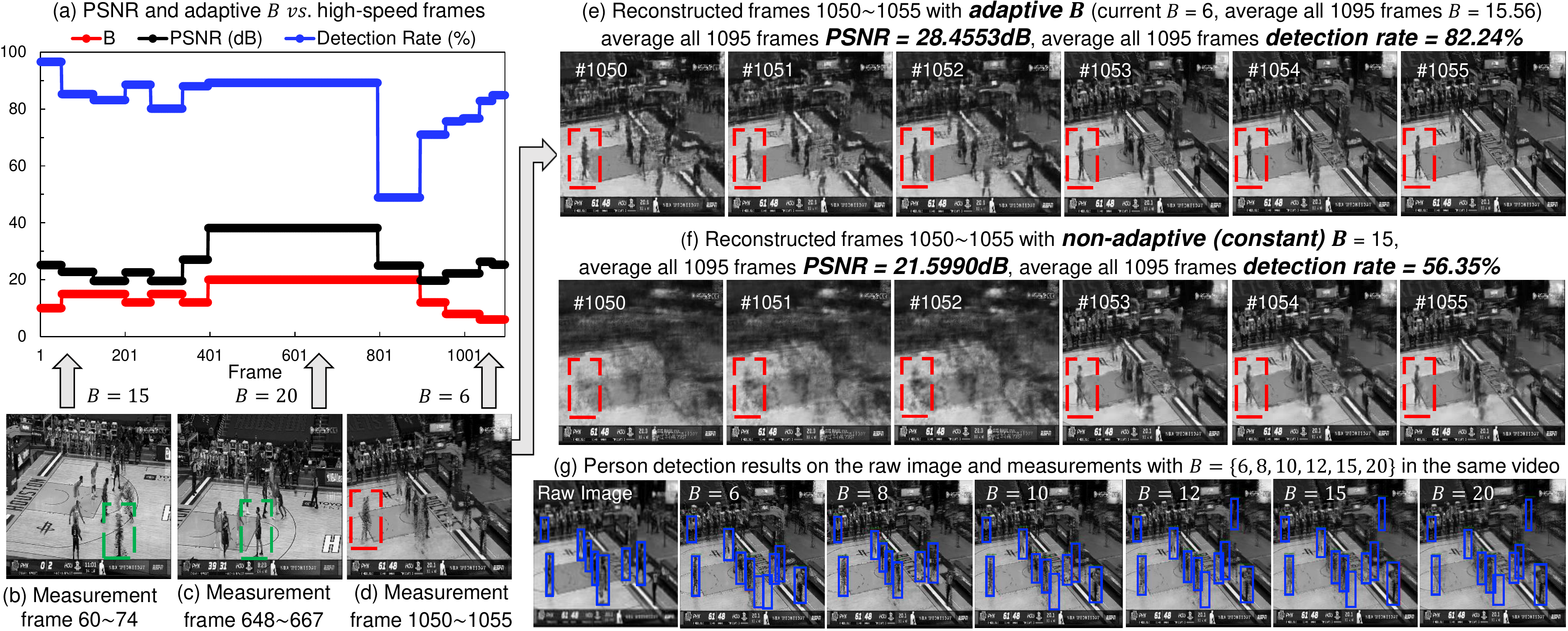}
\caption{Adaptive $B$ from the detection rate on the measurements directly. (a) Reconstruction PSNR (dB) and adaptive $B$ (frames) (average $B$=15.56) from the {\em measurements}, against frame number. (b-d) Normalized measurements when the basketball players are running from the left-hand scene to stop at the right-hand scene, adapted $B$ = 15, 20, 6, respectively. (e) Reconstructed frames 1050$\sim$1055 from the measurement in (d) with {\em adaptive} $B$. (f) Reconstructed frames 1050$\sim$1055 with non-adaptive (constant) $B$=15. (g) Person detection results on the raw images and measurements with different $B = \{6, 8, 10, 12, 15, 20\}$ in the same video clip. Videos in the SM.}
\label{fig:nbaDetectionBB}
\end{figure*}

\subsection{RL based on Detection Rate}

In real life applications, the detection rates are sent to the RL module to adjust $B$. As mentioned before, we employ YOLOv3 \cite{redmon2018yolov3} on the original video dataset to obtain labels (bounding boxes of targets) and treat these labels as the ground truth. Then, we employ YOLOv3-Tiny~\cite{huang2018yolo}, a light-weight DL algorithm designed for resource-constrained devices, {\em on the measurements} to detect vehicles and person for the sake of speed.
The detection can also be performed on the {\em reconstructed videos}, which can potentially increase the accuracy by trading off power and latency~\cite{Lu20SEC}. 
In this work, aiming to conduct adaptive video CS on the end-user cases with limited power but requiring instant responses such as in self-driving vehicles, we use the {\em detection on measurements directly}.

In terms of detection metrics, a common way is to compute the intersection-over-union (IoU) between ground truth and prediction. IOU is a measure of the degree of overlap between two detected frames for target detection:
\begin{equation}
    {\rm IOU}=\frac{\operatorname{area}\left(BBOX_{p} \cap BBOX_{g t}\right)}{\operatorname{area}\left(BBOX_{p} \cup BBOX_{g t}\right)},
\end{equation}
where $BBOX_{gt}$ represents the bounding box of the ground truth (GT), and $BBOX_p$ of the predicted frame.
Predictions whose IoUs are larger than 0.5 are considered as true positives (TP). We use mAP (mean Average Precision) as our detection rate score: \begin{eqnarray}
\text{Precision} &=&\frac{TP}{TP+FP}=\frac{TP}{\text {all detections }}, \\
\text{Recall} &=&\frac{T P}{T P+F N}=\frac{T P}{\text {all ground truths }}, 
\end{eqnarray}
where $TP$ is the number of detection frames with IoU $>$ 0.5 and $FP$ with IoU $\leqslant $ 0.5 detection frames, or the number of redundant detection frames detecting the same GT. $FN$ refers to the number of missing detections.


In our four datasets, we only detect \texttt{vehicles} in the highway and urban scenarios, and in the other two scenarios, we only detect \texttt{persons}.

During implementation, we calculate the mAP for each batch size corresponding to $Q$ = $BatchSize\times B$ video frames (for the $BatchSize$ measurements). The reason for this is that the calculated {\texttt{DetectionRate}} (mAP)  will not fluctuate sharply, but will change with the scene within a certain range. This is also the adaptation time of our RL module and the $BatchSize$ can be set to one for fast adaptation in real applications. 
%
For the reward design, we set the threshold (lower bound) of the acceptable detection rate as 75\%, \ie, $drth$ = 75\%, and obey the reward mechanism in Algorithm~\ref{algo:RL_SCI} for adaptive video CS. We also show the PSNR of the reconstructed videos for comparison purposes.


{We believe that it is the right approach to compare our proposed method against a fixed compression ratio ($B$:1). For adaptive sensing of video CS considered here, the only paper related to ours is~\cite{Yuan13ICIP}, which considers the same problem by using a motion estimation method to adapt $B$. However, both the reconstruction algorithm and the adaptive sensing framework developed therein produce low-quality results. Specifically, it has been shown in~\cite{Qiao2020_APLP,Cheng20ECCV_Birnat} that the E2E-CNN used in this paper can provide much better results than the reconstruction algorithms used therein. 
Besides, the look-up table used therein is not flexible. Our main goal of this paper is to prove that RL works well in adaptive video compressed sensing.}

\vspace{2mm}
\noindent \textbf{Highway Scene:}
Figure~\ref{fig:highwayDetectionBB} presents the testing results based on the traffic video in the highway with the goal of detecting vehicles from the raw {\em adaptive measurements}. 
Specifically, Fig.~\ref{fig:highwayDetectionBB} (a) presents the changes in PSNR (dB), detection rate (\%) and {\em adaptive} $B$ (frames) from the measurements against a constant stream of traffic video frames. Starting from a random $B$, RL module adjusts $B$ based on the learned speed and content from the raw measurements. Similarly to Fig.~\ref{fig:urban},  we keep the original video speed of the first one-third of the video frames, then freeze the video for the middle, and finally skip every two frames to simulate a fast speed scenario for the last two one-third of video frames. 
Under the decision of our proposed RL, $B$ has approximately maintained a certain range at the beginning, then rises to a higher level ($B$ = 20 in the frozen frames), and then drops back to a lower level after a period of time (due to the high speed). Once a certain $B$ is decided, the calculated Detection Rate and PSNR will lead to the opposite change of $B$, \ie, an increased $B$ will lead to a decrease in the detection rate and PSNR, and vice versa.
Consequently, three normalized measurements with different values of adaptive $B$ are shown in  Fig.~\ref{fig:highwayDetectionBB} (b-d) with adaptive $B$ = 6, 20, 6. We can see that the normalized measurement (c) has the largest adaptive $B$ = 20 since its corresponding original video frames are stationary, while the normalized measurement (d) is blurry with the smallest adaptive $B$ = 6 due to the fast object speed in these video frames.   

This video has a total of 565 frames, achieving a mean compression ratio (average $B$) of 13.64.
To demonstrate the usability of adapting $B$ based on the sensed video data, we compare adaptive reconstructions (Fig.~\ref{fig:highwayDetectionBB}(e)) to those obtained when $B$ is fixed at or near its expected value (Fig.~\ref{fig:highwayDetectionBB}(f) at $B$=12). Fig.~\ref{fig:highwayDetectionBB}(f) shows the 
reconstructed frames 512$\sim$517 from the measurement in (d) with {\em non-adaptive} (constant) $B$. Comparing Figs.~\ref{fig:highwayDetectionBB}(e) and ~\ref{fig:highwayDetectionBB}(f), we notice that adapting $B$ provides a significant (4.3dB) higher reconstruction quality (average all 565 frames PSNR=26.37dB) than fixing $B$ even lower than its expected value (average PSNR=22.04dB). Besides, it also improves the average detection rate from 66.32\% to 75.51\%. To present the effects of diverse $B$ on the object detection based on measurements, we visualize the vehicle detection results on the raw (original) images and measurements with different $B$ = $\{6, 8, 10, 12, 15, 20\}$ in the same video clip in Fig.~\ref{fig:highwayDetectionBB}(g). It can be seen that a decent detection rate is obtained at $B$ = 6 or 8, while a larger $B$ will lead to false alarms.





\begin{figure*}[htbp!]
\centering
\includegraphics[width=\textwidth]{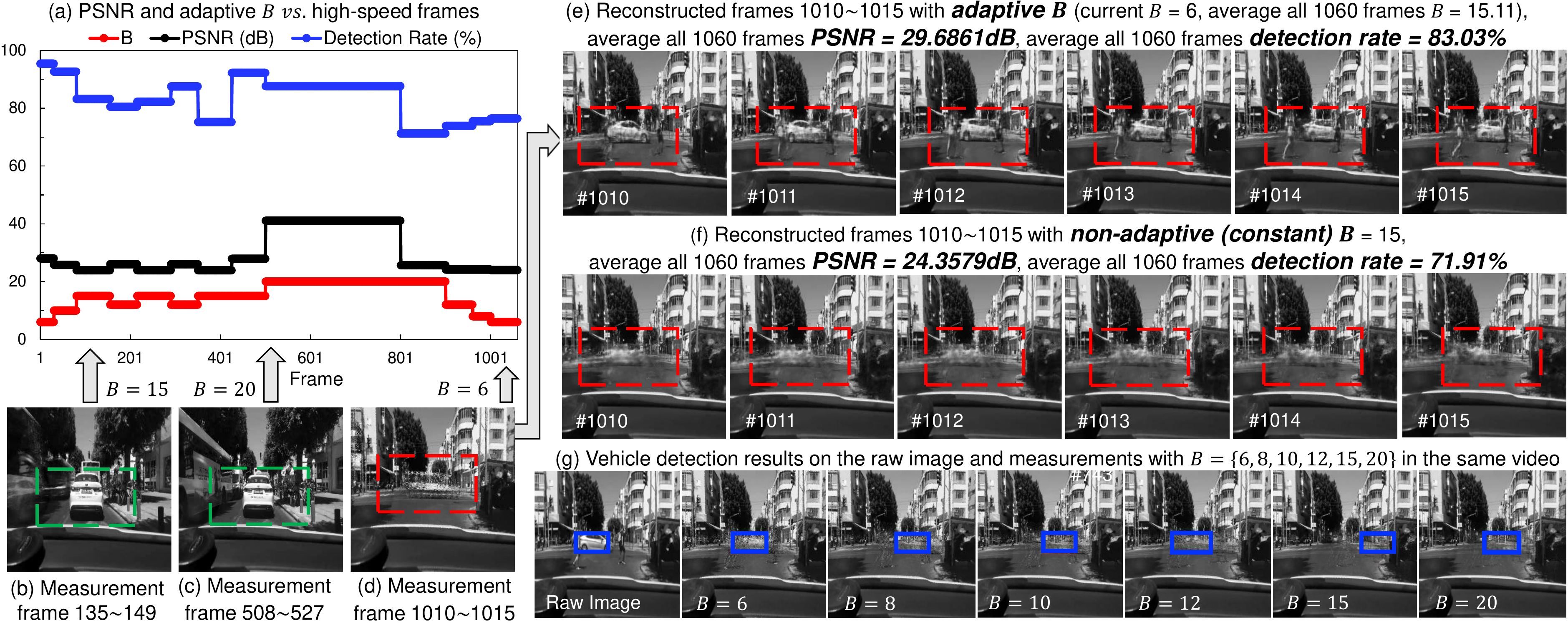}
\caption{Adaptive $B$ from based on the detection rate from the measurements directly. (a) Reconstruction PSNR (dB) and adaptive $B$ (frames) (average adaptive $B$ = 15.11) from the {\em measurements}, all are plotted against frame number. (b-d) Measurements when there is one moving front vehicle, one stopping front vehicle, and one front vehicle \textit{passing vertically and suddenly} inside the scene, adapted $B$ = 15, 20, 6, respectively. (e) Reconstructed frames 1010$\sim$1015 from the measurement in (d with {\em adaptive} $B$. (f) Reconstructed frames 1010$\sim$1015 with non-adaptive (constant) $B$ = 15. (g) Vehicle detection results on the raw images and measurements with different $B = \{6, 8, 10, 12, 15, 20\}$ in the same video clip.}
\label{fig:urbanDetectionBB}
\end{figure*}

\begin{figure*}[hb]
\centering
\includegraphics[width=\textwidth]{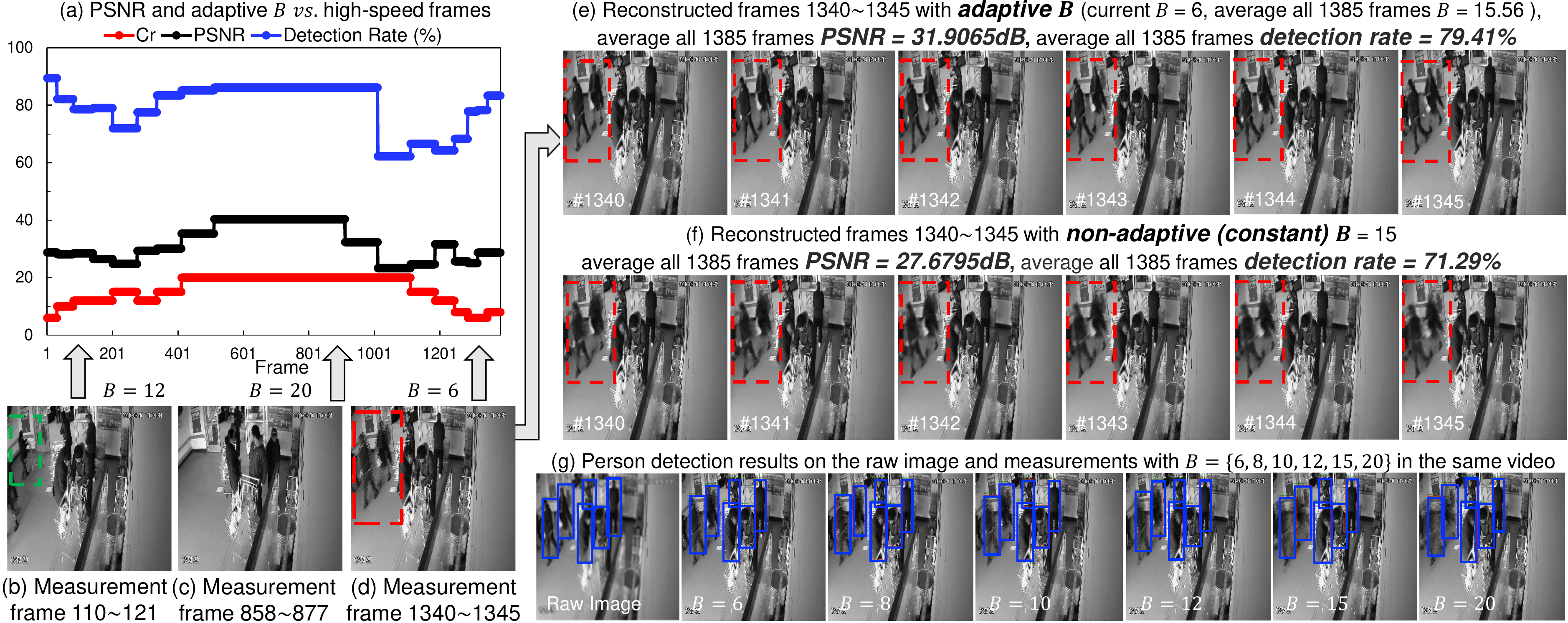}
\caption{Adaptive $B$ based on the detection rate from the measurements directly. (a) Reconstruction PSNR (dB) and adaptive $B$ (frames) (average adaptive $B$ = 15.84) from the {\em measurements}, all are plotted against frame number. (b-d) Measurements when there are one customer entering, no customers entering or leaving, and two customers leaving the grocery store, adapted $B$ = 12, 20, 6, respectively. (e) Reconstructed frames 1340$\sim$1345 from the measurement in (d with {\em adaptive} $B$. (f) Reconstructed frames 1340$\sim$1345 with non-adaptive (constant) $B$ = 10. (g) Person detection results on the raw images and measurements with different $B = \{6, 8, 10, 12, 15, 20\}$ in the same video clip.}
\label{fig:martDetectionBB}
\vspace{-0.6cm}
\end{figure*}

\vspace{2mm}
\noindent \textbf{NBA Scene:}
Following similar steps, Fig.~\ref{fig:nbaDetectionBB} presents the testing results for the publicly available NBA video. Unlike previous vehicle-related scenes, NBA videos are used to detect basketball players. Although the speed of human movement may be not as fast as that of vehicles, the corresponding inference of human-related video frames may not necessarily have better results. Because a single target (here is the person) occupies fewer pixels compared to vehicles, especially the rapid movement of players and mutual occlusion will make the measurements more blurry as in Fig.~\ref{fig:nbaDetectionBB}(b)-(d). As shown in Fig.~\ref{fig:nbaDetectionBB}(a), in the latter part, the detection rate has a relatively sharp drop, caused by the dramatic transition from slow to very rapid changes in adjacent frames of the video clip.
From the selected reconstructed frames in Fig.~\ref{fig:nbaDetectionBB}(e)-(f) and detection frames in (g), we can see that  adapting $B$ leads to a 6.85 dB improvement in PSNR and a 25.89\% increase in detection rate. 
This clearly verified the efficacy of our proposed RL for adaptive sensing in saving memory and bandwidth (an average higher $B$), power (detection on the raw measurements directly) and potential cost.





\vspace{2mm}
\noindent \textbf{Urban Scene:} Figure~\ref{fig:urbanDetectionBB} shows the testing result of an urban video clip taken by the front camera of a driving connected vehicle, with the goal of detecting surrounding vehicles from the raw {\em adaptive measurements}. Differently from the highway video, the captured surrounding vehicles have smaller relative speed compared with the camera (host vehicle) at the beginning, as the host and surrounding vehicles are driving along the same road. Then the traffic light at the intersection turns from green to red, and the relative speed differences between the host and surrounding vehicles become smaller and smaller until all vehicles become stationary. In the latter part of this video, the traffic lights become green again and all vehicles speed up aiming to cross the intersection. Here, we can notice some front vehicles passing  perpendicularly with respect to the image plane with higher speed \textit{suddenly}, which \textit{simulates the driving situation where pedestrians or vehicles suddenly cross the road and the host vehicle needs a quick emergency response by analyzing captured measurements to avoid collisions and fatal crashes}.


Specifically, Fig.~\ref{fig:urbanDetectionBB} (a) presents the changes in reconstruction PSNR (dB), detection rate (\%) and the related {\em adaptive} $B$ (frames) from the measurements against a constant stream of traffic video frames. Starting from a random $B$, the RL module adjusts $B$ based on learning the speed and content from the raw measurements. Three measurements with different values of the adaptive $B$ are shown in Fig.~\ref{fig:urbanDetectionBB} (b-d) with adaptive $B$ = 15, 20, 6. We can see that the measurement is clear with the largest adaptive $B$ = 20 since its corresponding original video frames are stationary, while measurement (d) is more blurry with the smallest adaptive $B$ = 6 due to the fast speed of the related video frames and the fast speed of the front vehicle that is passing perpendicularly to the camera.   
This video takes a total of 1060 frames to capture, achieving a mean compression ratio (average $B$) of 15.11.

To demonstrate the usability of adapting $B$ based on the sensed video data, we compare adaptive reconstructions (Fig.~\ref{fig:urbanDetectionBB}(e)) to those obtained when $B$ is fixed at or near its expected value (Fig.~\ref{fig:urbanDetectionBB}(f) at $B$=15). Fig.~\ref{fig:urbanDetectionBB}(f) shows the 
reconstructed frames 1010$\sim$1015 from the measurement in (d) with {\em non-adaptive} (constant) $B$. Comparing Fig.~\ref{fig:urbanDetectionBB}(e) and Fig.~\ref{fig:urbanDetectionBB}(f), we notice that adapting $B$ provides a significant (5.3dB) higher reconstruction quality (average PSNR of all 1060 frames is equal to 29.69dB) than fixing $B$ even lower than its expected value (average PSNR=24.36dB). Besides, it also improves the average detection rate from 71.91\% to 83.03\%. To present the effects of diverse $B$ on the object detection based on measurements, we visualize the vehicle detection results on the raw images and measurements with different $B$ = $\{6, 8, 10, 12, 15, 20\}$ in the same video clip in Fig.~\ref{fig:urbanDetectionBB}(g).

\vspace{2mm}
\noindent \textbf{Grocery Store Scene:}
Following similar steps, Fig.~\ref{fig:martDetectionBB} presents the testing results based on the surveillance videos collected from a middle-sized grocery store. As shown in Fig.~\ref{fig:martDetectionBB}(a), $B$ has approximately maintained a certain range at the beginning, then rises to a higher level ($B$ = 20 in the frozen frames), and then drops back to a lower level after a period of time (due to high speed). Once a certain $B$ is decided, the calculated detection rate and PSNR will lead to the opposite change of $B$, \ie, an increased $B$ will lead to a decrease in the detection rate and PSNR, and vice versa. From the exemplar reconstruction frames in Fig.~\ref{fig:martDetectionBB}(e)-(f) and detection frames in (g), we can see that our adaptive $B$ provides a higher (4.2dB) reconstruction quality than fixing $B$ even lower than its expected value, and it also improves the average detection rate from 71.29\% to 79.41\%.


\subsection{Performance of the Reconstruction}

\begin{figure}[h]
\centering
\includegraphics[width = \columnwidth]{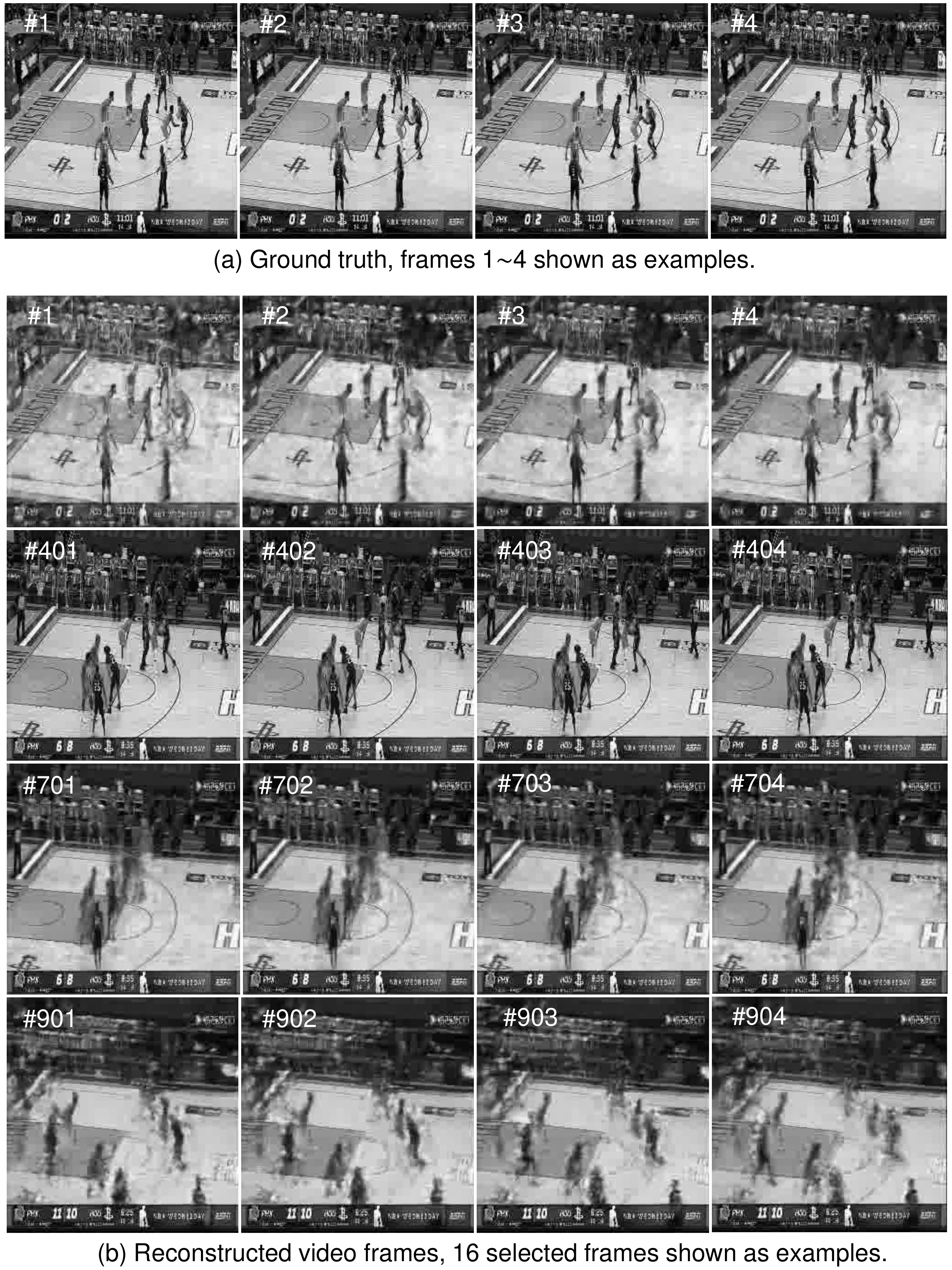}
\caption{Selected reconstructed frames (b) based on the adaptive $B$ presented in the NBA scene. Frames 1 to 4 in (a) are shown as examples of ground truth.}
\label{fig:nbaTruth}
\end{figure}

\noindent \textbf{Person Related Videos:}
Figure~\ref{fig:nbaTruth} implements an adaptive $B$ on the NBA video.  Fig.~\ref{fig:nbaTruth}(a)  presents the ground truth of the first four frames as examples. Several reconstructed frames based on the adaptive $B$ are shown in Fig.~\ref{fig:nbaTruth}(b). In comparison, the reconstructed images of the NBA video are more blurry than those in the grocery store video since the movement speed of players is much higher than the speed of customers.

\begin{figure}[h]
\centering
\includegraphics[width=\columnwidth]{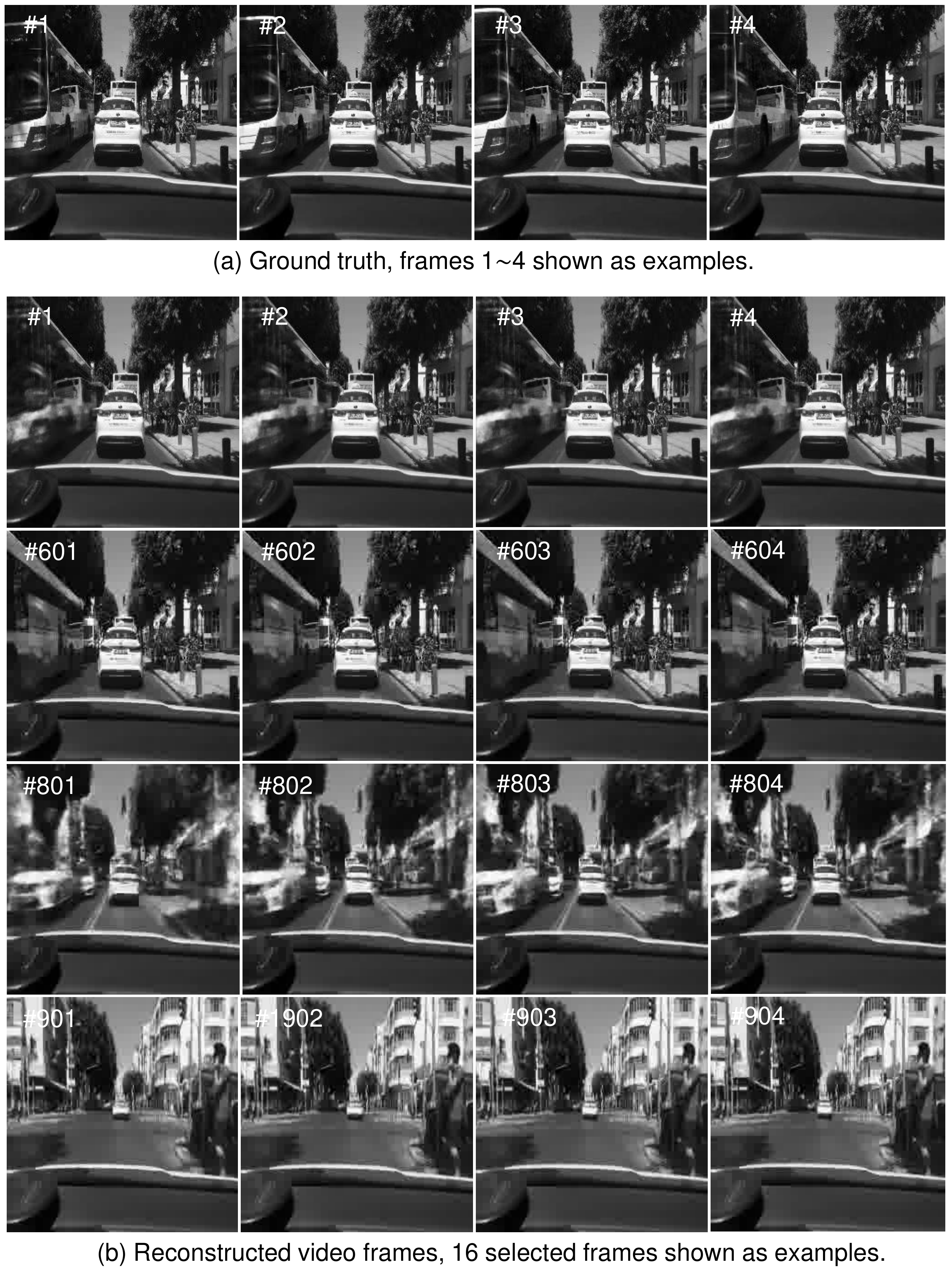}
\caption{Selected reconstructed frames (b) based on the adaptive $B$ presented in the urban scene. Frames 1 to 4 in (a) are shown as examples of ground truth.}
\label{fig:urbanTruth}
\vspace{-0.4cm}
\end{figure}

\begin{figure}[h]
\centering
\includegraphics[width = \columnwidth]{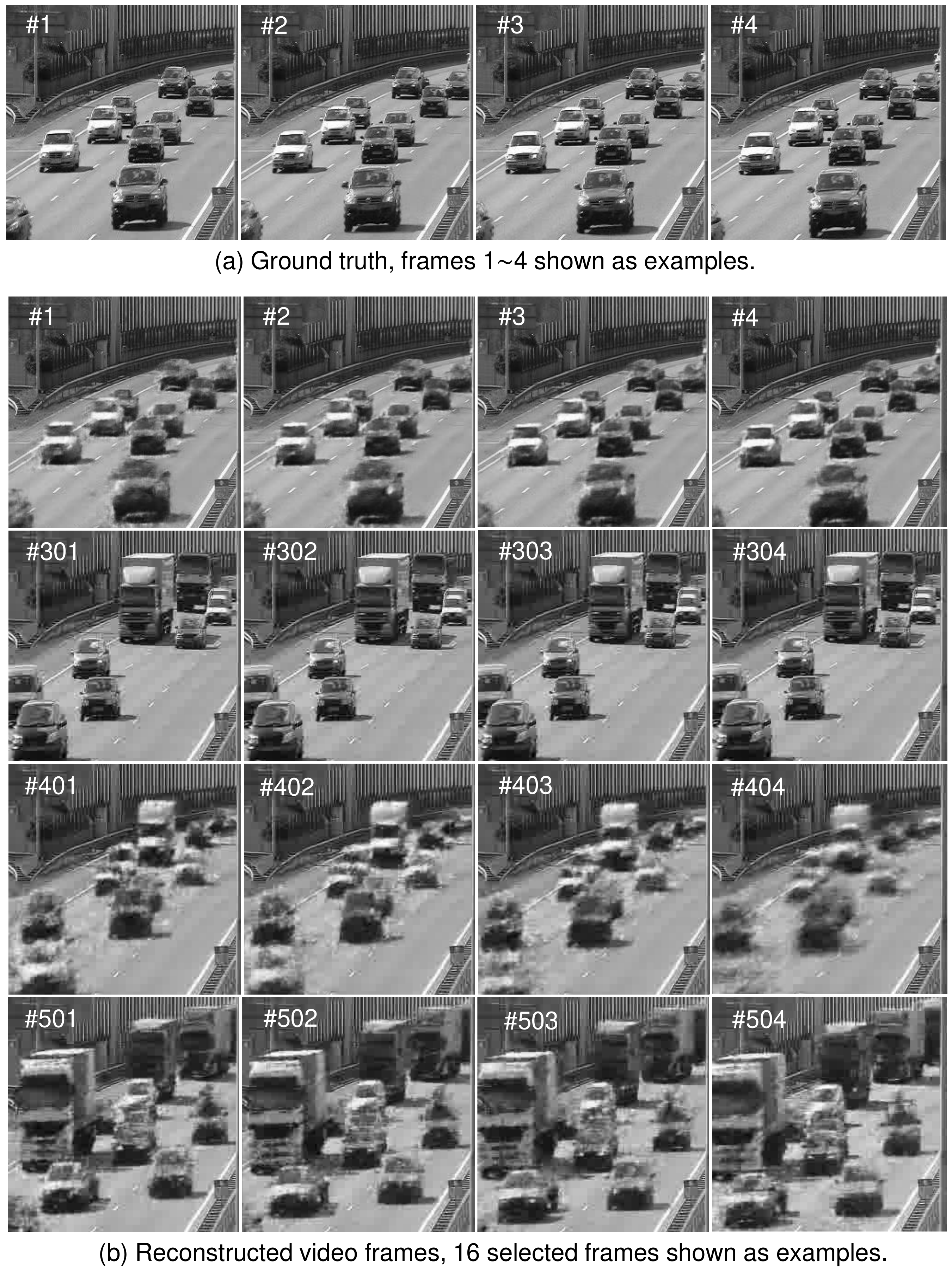}
\caption{Selected reconstructed frames (b) based on the adaptive $B$ presented in the highway. Frames 1 to 4 in (a) are shown as examples of ground truth.}
\label{fig:highwayTruth}

\end{figure}

\vspace{2mm}
\noindent \textbf{Vehicle Related Videos:}
Similarly, Fig.~\ref{fig:urbanTruth} and Fig.~\ref{fig:highwayTruth} implement adaptive $B$ on the urban video and the highway video captured by the front camera of a driving vehicle and the traffic camera, respectively.  Fig.~\ref{fig:urbanTruth}(a) and Fig.~\ref{fig:highwayTruth}(a) also present the ground truth of the first four frames as examples. Selected reconstructed frames based on the adaptive $B$ are presented in Fig.~\ref{fig:urbanTruth}(b) and Fig.~\ref{fig:highwayTruth}(b). 


It can be seen from these plots that by using our proposed adaptive video sensing approach, the reconstructed frames are consistently at a high quality level.

\subsection{Additional Considerations}

\noindent \textbf{Recovery from Noisy Measurements:}
We also verified the proposed RL module's robustness to noise by investigating the recovery from noisy measurements. Specifically, as shown in Table~\ref{Tab:noise}, when zero-mean Gaussian noise $\nv\sim{\cal N}(0,\sigma)$ is added to the measurements (normalized to $[0,1]$), both the quality of the reconstruction (as measured by PSNR in dB), as well as the detection rates (DR, 1 is the highest value) are high for different noise levels. 

\begin{table}[htbp!]
\caption{Reconstruction PSNR and detection rate \textit{vs.} noise $\sigma$.}
\centering
\footnotesize
\scalebox{1}{
\begin{tabular}{|c|c|c|c|}
\hline
\diagbox{$\sigma$}{PSNR, DR}{$B$}  & \textbf{6}    & \textbf{10}   & \textbf{15}   \\ \hline
\textbf{0}     & 28.73, 0.8543 & 28.44, 0.8557 & 28.33, 0.8138 \\ \hline
\textbf{0.005} & 28.56, 0.8521 & 28.30, 0.8436 & 28.19, 0.8018 \\ \hline
\textbf{0.010}  & 28.18, 0.8374 & 27.99, 0.8162 & 27.89, 0.7745 \\ \hline
\textbf{0.050}  & 24.70, 0.7534 & 24.62, 0.7633 & 24.52, 0.7126 \\ \hline
\textbf{0.100}   & 21.58, 0.7147 & 21.52, 0.7123 & 21.44, 0.6849 \\ \hline
\end{tabular}}
\label{Tab:noise}
\end{table}

\vspace{2mm}
\noindent \textbf{Inference Speed:}
In addition, the inference speed of our RL module is high for many time-sensitive applications. For example, in terms of autonomous driving, when a connected and autonomous vehicle (CAV) is driving in an urban area at a speed of 40 kilometers per hour, the execution time of each real-time task should be less than 100 milliseconds \cite{lu2021VC}. On average, our whole RL module for inference takes 12 milliseconds per measurement. The inference time of object detection models, \ie, YOLOv3 and YOLOv3-Tiny, are 42 milliseconds and 16 milliseconds, respectively. Regarding the E2E-CNN (not necessary), the inference time is 29 milliseconds. The total of all those inference speeds is much less than 100 milliseconds, which shows actionable insights of employing our work for real-world CAV applications.

\vspace{2mm}
\noindent \textbf{Practicality to Real Systems:}
Moreover, recent advances in reconstruction networks have resulted in excellent results by { training on simulated data} in an offline manner \cite{Qiao2020_APLP}. Hence, we opine that with the RL model, {\em training on simulated data and performing inference on real data} will work as well.

\section{Conclusions \label{Sec:conclusion}}
We introduce reinforcement learning to perform adaptive temporal compressive sensing of video.
The proposed RL algorithm conducts adaptive sensing directly on the raw measurements and thus saves memory, bandwidth and power on the end-users equipped with SCI cameras. Extensive results demonstrated the potential of our proposed methods in real life applications of video compressive sensing. 
We are working on building an end-to-end system of video SCI and RL to conduct real-time adaptive sensing experiments and demonstrations using our proposed algorithm.

{\footnotesize
\bibliographystyle{IEEEtran}
\bibliography{main}
}












\end{document}